%% file: main.tex
\documentclass[letterpaper, 10 pt, conference]{ieeeconf}
\IEEEoverridecommandlockouts   %
\makeatletter
\def\section{\@startsection{section}{1}{\z@}{1.0ex plus 1ex minus 0.5ex}%
{0.4ex plus 0.6ex minus 0ex}{\normalfont\normalsize\centering\scshape}}%
\def\subsection{\@startsection{subsection}{2}{\z@}{1.0ex plus 1ex minus 0.5ex}%
{0.4ex plus 0.3ex minus 0ex}{\normalfont\normalsize\itshape}}%
\skip\footins 0.5\baselineskip plus 0.2\baselineskip minus 0.2\baselineskip
{\footnotesize\global\footnotesep 0.72\baselineskip}
\makeatother

\usepackage{amsmath,amssymb}
\usepackage{graphicx}
\usepackage{array}
\usepackage{booktabs}
\usepackage{xcolor}
\usepackage{cite}
\usepackage{url}
\usepackage{tikz}                  %
\usetikzlibrary{arrows.meta,positioning,fit,calc,backgrounds,decorations.pathreplacing}
\definecolor{cFull}{HTML}{0072B2}  %
\definecolor{cVision}{HTML}{E69F00}
\definecolor{cLight}{HTML}{EFEFEF}

\newcommand{\vect}[1]{\mathbf{#1}}
\newcommand{\R}{\vect{R}}

\makeatletter
\def\ps@ieeenotice{%
  \let\@mkboth\@gobbletwo
  \def\@oddhead{}\def\@evenhead{}%
  \def\@oddfoot{\parbox[t]{\textwidth}{\centering\fontsize{7}{8.2}\selectfont
    This work has been submitted to the IEEE for possible publication. Copyright may be
    transferred without notice, after which this version may no longer be accessible.\par}}%
  \let\@evenfoot\@oddfoot}
\makeatother

\begin{document}

\title{\LARGE \bf
Recording Hand-Held Laparoscopic Instrument Motion in the Operating Room:
Magnetometer-Free Fusion of Inertial, Range and Visual Sensing}

\author{Jiyul Lee$^{*,1,2,3}$, Dongho Yee$^{1,2,4,5}$, Juahn Oh$^{1,2,8}$, Jinseok Lee$^{2,4}$,
Yechan Seo$^{1,2,3}$,\\
Seong Jeong$^{1,2,3}$, Minsung Kim$^{1,2,4}$, Seonho Shim$^{2,9}$, Younghoon Noh$^{2,4}$,
Hyuk Choi$^{1,2,3}$,\\
Youngbin Kong$^{1,7}$ and Hyoun-Joong Kong$^{\dagger,1,3,6}$%
\thanks{$^{*}$First author. $^{\dagger}$Corresponding author.}%
\thanks{$^{1}$Department of Transdisciplinary Medicine, Seoul National University Hospital,
Seoul, Republic of Korea. $^{2}$Rosota Inc., Seoul, Republic of Korea. $^{3}$Department of
Medicine, Seoul National University College of Medicine, Seoul, Republic of Korea.
$^{4}$Department of Mechanical Engineering, Seoul National University, Seoul, Republic of
Korea. $^{5}$Department of Computer Science and Engineering, Seoul National University, Seoul,
Republic of Korea. $^{6}$Institute of Convergence Medicine with Innovative Technology, Seoul
National University Hospital, Seoul, Republic of Korea. $^{7}$Interdisciplinary Program in
Medical Informatics, Seoul National University College of Medicine, Seoul, Republic of Korea.
$^{8}$Eulji University College of Medicine, Daejeon, Republic of Korea. $^{9}$Department of
Mechanical Engineering, Chungang University, Seoul, Republic of Korea.}}

\maketitle
\thispagestyle{ieeenotice}   %
\pagestyle{empty}

\begin{abstract}
Most minimally invasive procedures are still performed with hand-held
laparoscopic instruments, yet only the endoscopic video is retained; the
instrument motion that expresses surgical skill, and that could support skill
assessment and robot learning, is lost. Pose from video alone remains
millimeters to centimeters off, and an instrument-mounted inertial measurement
unit (IMU) cannot rely on its magnetometer, whose field changed with tool pose
and between sessions in our measurements.
 We present a surgical
instrument-state logger that clips onto a conventional instrument without
modifying the part that enters the patient and fuses a six-axis IMU and a
time-of-flight (ToF) rangefinder with a markerless camera in an error-state
Kalman filter under the remote center of motion (RCM) of the trocar. Heading
comes from the shaft silhouette, segmented by a U-Net, in place of the
magnetometer: the rotation-angle error is 0.200$^\circ$, against 3.58$^\circ$
from the accelerometer and magnetometer alone. Against a Franka Research~3
manipulator, and without alignment to it, the displacement error over 300
translation trials was 1.21\,mm RMS and the relative-rotation error over 180
rotation trials 0.34$^\circ$ RMS. On continuous trajectories, tracked and
displayed in real time, the absolute tip error was 1.22\,mm (programmed) and
3.04\,mm (teleoperated) after post-hoc tuning of three filter parameters, and
the full fusion beat every sensor subset. Because the estimator uses no magnetic measurement, its accuracy does not rely
on an undisturbed field. The same clip-on device could thus record metric tip
trajectories during routine hand-held laparoscopy, while displaying the
insertion depth and attitude that are hidden once the instrument is inside the
patient.
\end{abstract}

\begin{figure*}[tb]
\centering
\includegraphics[width=0.90\textwidth]{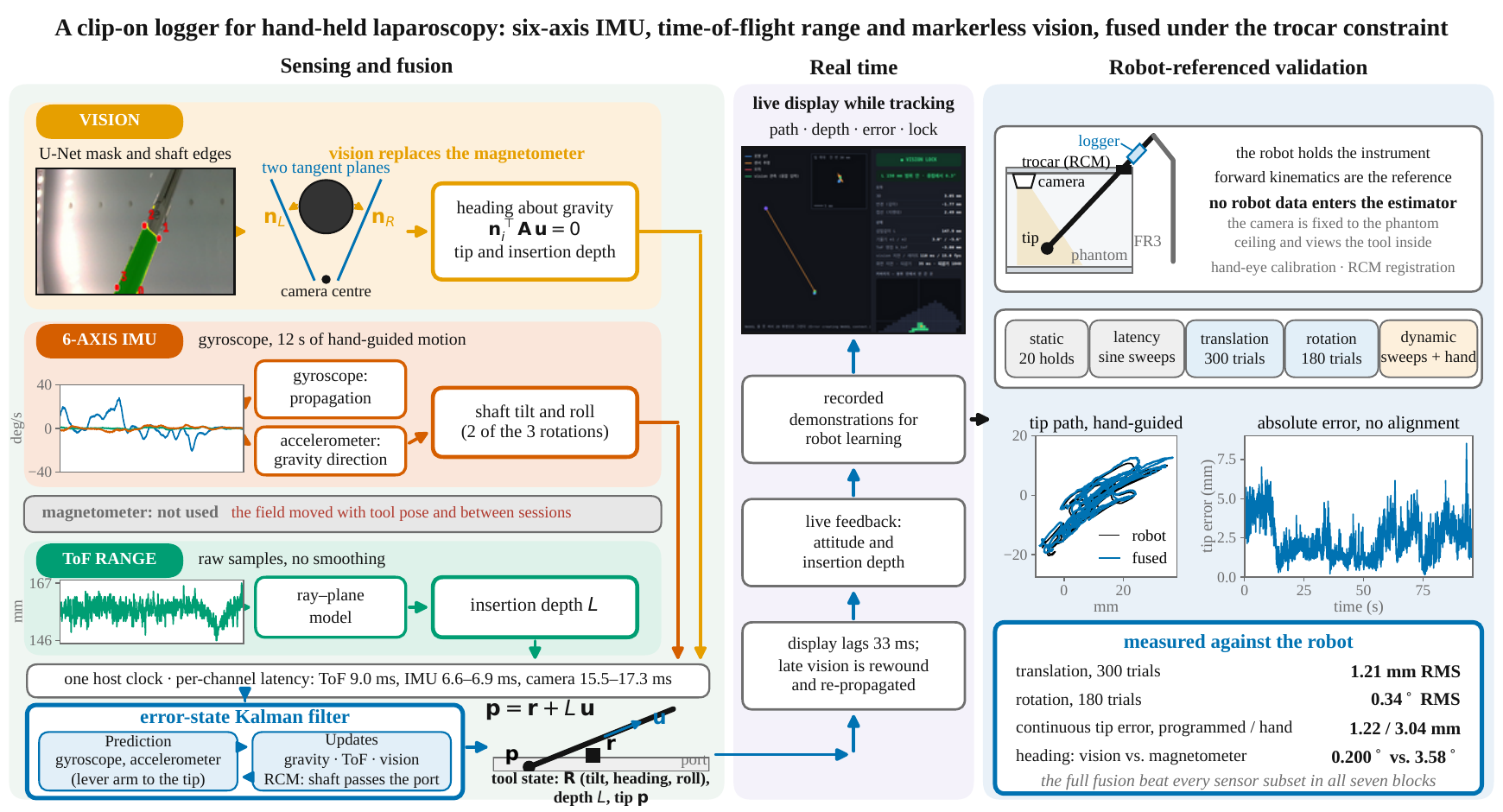}
\caption{Overview of the logger. \emph{Left:} the sensing lanes and the error-state Kalman filter
under the trocar constraint. \emph{Center:} the live display. \emph{Right:} the robot-referenced
protocol and the measured accuracy.}
\label{fig:hero}
\end{figure*}

\section{Introduction}

Most minimally invasive surgery is still performed by hand with conventional
laparoscopic instruments: robotic use across common general-surgery procedures
reached only 15.1\% by 2018~\cite{sheetz2020trends}, and robotic assistance
accounted for 5.2\% of Medicare cholecystectomies in
2019~\cite{kalata2023comparative}. A robotic system logs the kinematics of
every case it performs, whereas a hand-held instrument logs nothing. A metric
record of how the instrument moved would serve surgical data science and skill
assessment~\cite{maierhein2022sds,lam2022skill}, and would supply
demonstrations for surgical robot
learning~\cite{kim2024srt,kim2025srth,schmidgall2024foundation}, which today
are collected on the robot itself.

The endoscopic video is the natural source, but monocular pose estimates remain
millimeters to centimeters off, worst along the viewing
ray~\cite{allan2018pose,chen2025surgipose,shabir2025markerless}.

We instead measure the motion on the instrument itself, with a module clipped
to the handle side of a standard instrument so that the part entering the
patient stays unmodified. An IMU on the instrument, however, obtains heading about gravity only from its
magnetometer, and we met the known distortion of magnetic fields around
operating equipment in our first prototype: the measured field changed with tool
pose and between sessions, and the on-chip nine-axis heading drifted by up to
2.66$^\circ$ within 30\,s at rest (Sec.~\ref{sec:six}).

We removed the magnetometer from the estimate and let vision take its place.
The resulting surgical instrument-state logger fuses a six-axis IMU and a ToF
rangefinder on the instrument with a camera that views the field (fixed above a
phantom here; the endoscope in clinical use) under the RCM of the trocar
(Fig.~\ref{fig:hero}):
gravity fixes tilt and roll, the shaft silhouette fixes heading, and the
rangefinder fixes insertion depth. Against a Franka Research~3 (FR3) robot
holding the instrument, it tracked continuous motion with 1.22--3.04\,mm
absolute tip error, well below the video-only errors above although under
different conditions (Table~\ref{tab:prior}), while showing attitude and
insertion depth live, with neither markers nor a change to the instrument. The
error-state filter, the fusion of inertial, range and visual sensing and the RCM
constraint are each established; what we contribute is the design decision that
follows from the magnetic measurements above, and a validation strict enough to
test it:
\begin{enumerate}
\item \textbf{Vision in place of the magnetometer}, argued from measurement: the
magnetometer cannot supply heading on a surgical bench, and the shaft silhouette
observes the same degree of freedom about gravity without a marker or a change
to the instrument.
\item \textbf{An estimator built around that choice}, fusing a six-axis IMU, raw
ToF range and markerless vision under the RCM constraint of the trocar, running
live on a module clipped to a standard instrument.
\item \textbf{Robot-referenced, alignment-free validation} of stability,
latency, translation, rotation and continuous tracking against a moving robot,
with a per-sensor ablation in every experiment that shows which degree of
freedom each sensor holds.
\end{enumerate}

\section{Related Work}
\label{sec:related}

\begin{table}[tb]
\caption{How prior instrument tracking was validated}
\label{tab:prior}
\centering\footnotesize\setlength{\tabcolsep}{2pt}
\begin{tabular}{@{}llll@{}}
\toprule
Work & Sensing & Reference & Reported error \\
\midrule
Heiliger~\cite{heiliger2023tracking} & IMU on handle & phantom & ${\le}6.4^\circ$, no position \\
Ebina~\cite{ebina2024motion} & IR markers & 2nd tracker & 2.6--4.4\,mm, 0.9--1.4$^\circ$ \\
Gautier~\cite{gautier2021realtime} & tape markers & robot & max.\ 1.5--4\,mm \\
Shabir~\cite{shabir2025markerless} & video & marker & 3.8\,mm ($z$), 5.7$^\circ$ \\
SurgiPose~\cite{chen2025surgipose} & video, dVRK & robot & 9.7--12.0\,mm \\
\midrule
Ours & IMU+ToF+video & robot, moving & 1.22/3.04\,mm, 0.6--2.2$^\circ$ \\
\bottomrule
\end{tabular}
\\[1pt]\parbox{\columnwidth}{\footnotesize Errors as each work defines them, under different conditions. Ours: absolute RMS
tip error (programmed/teleoperated) and attitude error per block, against a
moving robot.}
\end{table}

\subsection{Vision-Based Instrument Pose Estimation}
Endoscopic video is the one signal that is already recorded, and a large body of
work estimates instrument pose from it. Articulated models fitted to endoscopic
images localize an instrument well within the image plane but degrade along the
viewing ray~\cite{allan2018pose}, and occlusion by tissue or smoke removes the
features the fit relies on~\cite{xu2024occlusion}. Learned estimators report
9.7--12.0\,mm displacement error on da~Vinci
trajectories~\cite{chen2025surgipose}, and a markerless method for a manual
instrument in a box trainer 3.8\,mm in depth with 5.7$^\circ$ in shaft
orientation~\cite{shabir2025markerless}. Unlike these estimators, we do not ask
vision for the full pose, only for what it conditions well: heading about
gravity and the depth of the shaft junction.

\subsection{Inertial and Hybrid Instrument Tracking}
An instrument-mounted IMU does not depend on line of sight. Heiliger
et~al.\ clip one to the handle and report at most 6.4$^\circ$ of orientation
error against a printed phantom, with no position~\cite{heiliger2023tracking}.
Optical markers add position: an infrared system reports 2.6--4.4\,mm against a
second tracker~\cite{ebina2024motion} and taped markers 1.5--4\,mm maximum
error against a robot~\cite{gautier2021realtime}, while visual--inertial
localization under a trocar constraint has been demonstrated on a hand-held
laparoscope~\cite{hartwig2022constrained}. Nine-axis modules share one
dependence: heading about gravity comes from the magnetometer, and magnetic
tracking near operating equipment is
distorted~\cite{franz2014em,sorriento2020tracking}. Hard- and soft-iron
calibration compensates only material fixed to the
sensor~\cite{kok2017inertial}, and a single metal plate has been reported to
raise magnetometer heading error from 3.3$^\circ$ to
17.0$^\circ$~\cite{chen2017imu}. Unlike prior instrument-mounted IMUs, our
estimator uses no magnetic measurement.

\subsection{Demonstration Capture and Kinematic Constraints}
Outside surgery, hand-held rigs capture manipulation demonstrations without a
robot: UMI~\cite{chi2024umi} and DexCap~\cite{wang2024dexcap} combine a hand-held
gripper or glove with cameras and inertial sensing to train manipulation
policies. The trocar acts as an RCM, a constraint used
to condition camera and instrument pose estimation and
control~\cite{vasconcelos2018rcm,gruijthuijsen2018fulcrum}. We adopt both ideas
for laparoscopy and carry the RCM inside the estimator. Table~\ref{tab:prior}
lists the reference each system above was validated against; unlike prior work,
we validate a marker-free, clip-on sensor set against a robot along continuous
trajectories, report absolute error without trajectory alignment, and ablate
every sensor on the same recordings.

\section{Methods}
\label{sec:methods}
This section covers the device and the trocar geometry, the preliminary study
that rules the magnetometer out, the vision and range models, and the filter
that fuses them. Table~\ref{tab:roles} lists which sensor fixes which
degree of freedom. Their weaknesses are complementary: the gyroscope drifts,
gravity is corrupted by motion, vision is intermittent and late, and ToF must be
interpreted through the orientation.

\begin{figure}[tb]
\centering
\includegraphics[width=0.94\columnwidth]{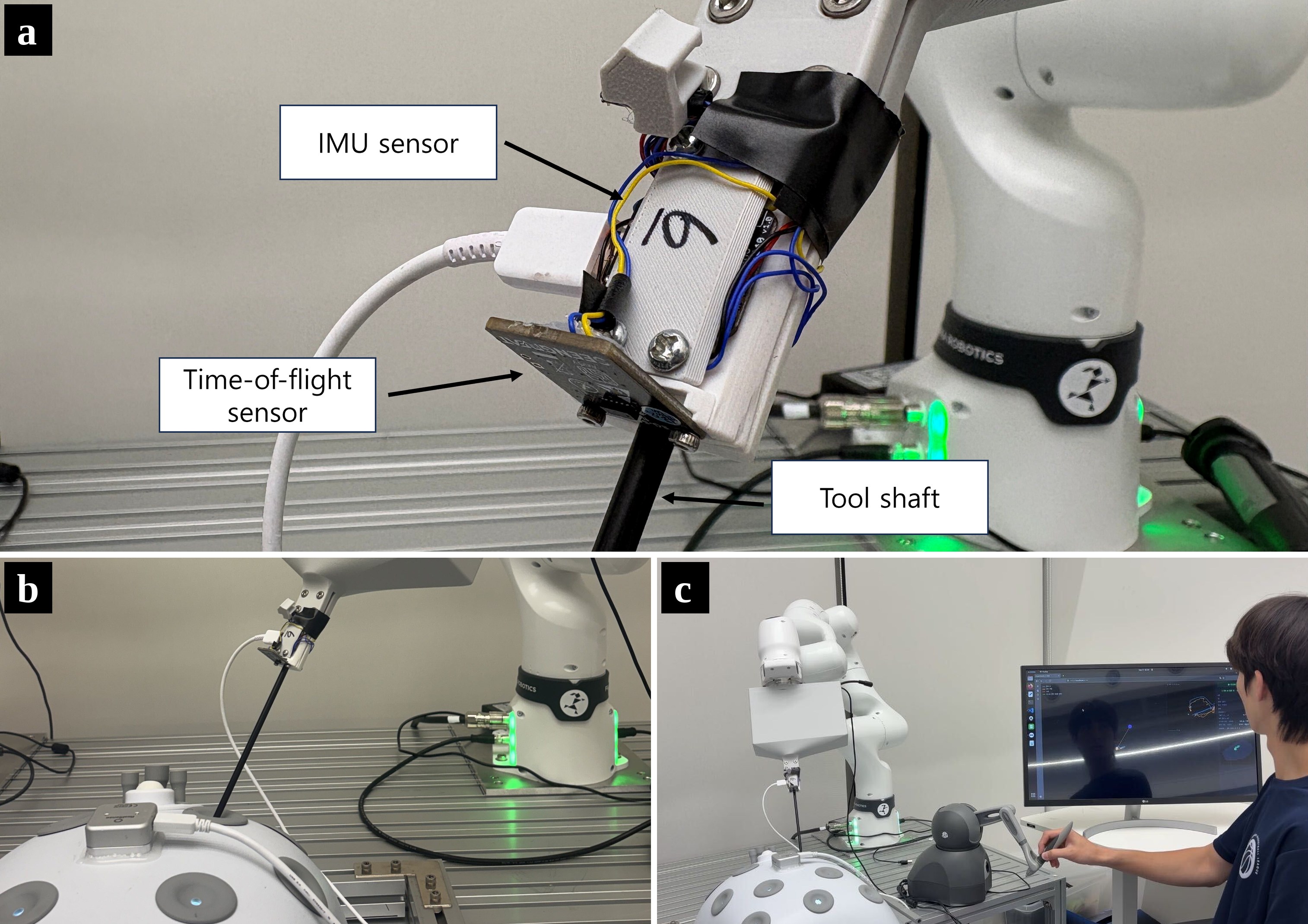}
\caption{The logger and the bench. (a) The clip-on module on the handle side of a standard
instrument. (b) The FR3 holds it through the trocar of the phantom, the camera above the field.
(c) Teleoperated recording with a haptic stylus and the live display.}
\label{fig:overview}
\end{figure}

\subsection{Hardware and Implementation}
\label{sec:hw}
The module clamps to the handle side of a standard laparoscopic instrument of
5.18\,mm shaft diameter (Fig.~\ref{fig:overview}a). It carries a BNO085 IMU, whose magnetometer is
recorded but never used, a VL53L0X ToF rangefinder whose beam points distally
toward the port, and an nRF52840 microcontroller that streams gyroscope,
accelerometer and raw range samples with their validity status at 200, 240 and
100\,Hz. An Intel RealSense D405 camera above the phantom records
$1280\times720$ color frames at 15\,fps with locked exposure, and every sample
is time-stamped on arrival on a common host clock. The U-Net is a ResNet-18 encoder--decoder trained on 1740
labeler-generated frames from a separate markerless session, split by pose with
25\% held out, using 512-px crops and 60 epochs of Adam at $3\times10^{-4}$, half of each batch
drawn from synthetic composites. Estimation, segmentation and the display run on one workstation (Intel
i7-13700K, RTX 4090).

\subsection{Geometry and Robot Control}
\label{sec:geom}
Let $W$ be the robot base, $\vect{r}\in\mathbb{R}^3$ the RCM at the port,
$\R\in SO(3)$ the tool orientation and $L$ the insertion depth. With
$\vect{e}_s$ the shaft axis in tool coordinates, the shaft direction is
$\vect{u}=\R\,\vect{e}_s$ and the tip is
\begin{equation}
\vect{p} = \vect{r} + L\,\vect{u}.
\label{eq:tip}
\end{equation}
The RCM anchors the shaft line, so the tip~\eqref{eq:tip} needs only the depth
$L$. The IMU and the ToF sensor sit 304\,mm and 284\,mm proximal to the tip. The
FR3 holds the instrument; its forward kinematics are the ground
truth~\cite{haddadin2022franka}, and the estimator uses sensor data and
calibration constants only.

Trials are commanded as tip positions in the camera frame and mapped to $W$ by
the hand-eye transform. The RCM $\vect{r}$ is registered once, with a 10\,cm
mark on the shaft at the port center, and each command is projected onto the
port: only the direction $\vect{u}_d$ from $\vect{r}$ and the depth, clipped to
$[L_{\min},L_{\max}]$, are retained, so a command that does not pass through the
port becomes a pivot about $\vect{r}$ rather than a lateral force on the
trocar. With the tip velocity
$\vect{v}=k_p(\vect{p}_d-\vect{p})+k_f\dot{\vect{p}}_d$ toward the projected
target $\vect{p}_d$, the commanded angular velocity is
\begin{equation}
\boldsymbol{\omega}=\frac{\vect{u}\times\vect{v}}{L}
  \;+\;k_{\mathrm{rcm}}\,\vect{u}\times\vect{u}_d,
\label{eq:rcmservo}
\end{equation}
whose first term pivots the shaft about $\vect{r}$ to move the tip and whose
second turns the shaft axis onto the port direction, servoing the shaft line
back through the port; roll is driven separately. Damped least squares on the
$6\times7$ tip Jacobian, with a posture term in its null space, converts the
twist into joint velocities at 1\,kHz. The same registered $\vect{r}$ serves as
the RCM in the estimator and in the ground-truth reference.

\begin{table}[tb]
\caption{The sensor suite: what each sensor measures, which degrees of freedom
it observes, and its measurement residual}
\label{tab:roles}
\centering\footnotesize\setlength{\tabcolsep}{3pt}
\input{tables/tab_sensors}
\end{table}

\subsection{Preliminary Study: Magnetic Disturbance and Accelerometer Drift}
\label{sec:six}
A nine-axis IMU takes heading from the magnetometer, which requires an
undisturbed field in every pose; two measurements before the main experiments
showed that this did not hold on our bench.

\emph{Pose dependence.} The Earth's field keeps a fixed angle to gravity, yet
across the 180 rotation trials of Sec.~\ref{sec:rotation} this angle changed
between the two poses of a trial by up to 4.8$^\circ$, against a 0.21$^\circ$
gravity residual. Rotation angles from the accelerometer and magnetometer alone
(TRIAD) erred by 3.58$^\circ$ RMS, 18 times the error of the proposed fusion.

\emph{Session dependence.} Between two static sessions recorded on one day the
mean field fell by 21\%, and the nine-axis heading drift within 30\,s holds rose
from at most 0.05$^\circ$ to 2.66$^\circ$ while accelerometer tilt drifted at
most 0.19$^\circ$.

\emph{Expected disturbance in the operating room.} Heading uses only the
horizontal field component $B_h=B\cos\delta$, so a horizontal disturbance
$\Delta_\perp$ rotates the heading estimate by $\arctan(\Delta_\perp/B_h)$ and
leaves tilt untouched: at $B_h\approx30\,\mu$T, 5\,$\mu$T already costs about
9.5$^\circ$.
Such a calibration covers neither the operating table nor the instruments being
exchanged~\cite{kok2017inertial}, and ten surgical facilities measured $64\pm20\,\mu$T on
their operating tables against a ${\sim}35\,\mu$T
background~\cite{nie2023subgauss}. The camera supplies heading instead. The
study also exposed a device effect: the BNO085 recalibrates its accelerometer at
run time without flagging it, and $\|\vect{a}\|$ jumped by 6.3\% within 13\,min
and settled 5.5\% above $g$. With that disabled in firmware, $\|\vect{a}\|$
stayed within 0.06\% of $g$.

\subsection{Markerless Shaft Geometry from Vision}
\label{sec:vision}

The D405 stereo depth is valid on only 8.5\% of the dark, smooth shaft pixels,
so depth must come from the color image. A rule-based labeler places each
silhouette edge at the gradient maximum in undistorted coordinates and fits a
trimmed line; on 300 frames of one fixed pose its axis-angle scatter was
0.013$^\circ$, against 0.409$^\circ$ for a human annotator. Its labels train a
U-Net~\cite{ronneberger2015unet}, and at run time both run on every frame, the
labeler supplying the lines and the network an independent coarse mask. A frame
reaches the filter only if the two masks overlap with an intersection over union
of at least 0.75 and their silhouette lines differ by less than 0.5$^\circ$ in
direction, 1.5\,px laterally and 4\,px at the shaft junction. No robot
data enters labeling or training.

Each silhouette line and the camera center span a plane tangent to the shaft
cylinder, with normals $\vect{n}_L,\vect{n}_R$. The axis
direction $\vect{n}_L\times\vect{n}_R$ is ill-conditioned because the planes
are only ${\sim}3.6^\circ$ apart (${\sim}8^\circ$ error), so we use the planes
as constraints instead,
\begin{equation}
\vect{n}_i^{\top}\vect{A}\,\vect{u}=0,\qquad i\in\{L,R\},
\label{eq:plane}
\end{equation}
with $\vect{A}$ the base-to-camera rotation, which reduces the residual to
0.35$^\circ$ on the same data.

The point of the axis closest to the camera follows from the \emph{sum} of the
normals and is well conditioned,
\begin{equation}
\vect{P}_0=\frac{\rho_s\,(\vect{n}_L+\vect{n}_R)}{1+\vect{n}_L^{\top}\vect{n}_R},
\label{eq:p0}
\end{equation}
where $\rho_s=2.59$\,mm is the measured shaft radius. Intersecting the back-projected ray of the shaft junction with the axis line
gives an insertion-depth observation; that intersection is ill-conditioned when
ray and axis are nearly parallel, so frames with
$1-(\vect{d}^{\top}\vect{u})^2<0.45$ are rejected rather than down-weighted.

\subsection{Time-of-Flight Range Model}
\label{sec:tof}
A single VL53L0X range scatters about 3\,mm (Table~\ref{tab:static}), so the
firmware smooths it with a five-sample moving average, at the cost of 39\,ms of
delay and of correlated outputs that a filter would count five times. The display uses the smoothed depth, the filter the raw
status-valid ranges.

An affine map $L=a\rho+c$ assumes the beam looks along the shaft onto a surface
normal to it; a cone model adds a fixed half-angle, so the range stretches with
tilt magnitude alone, blind to its direction. We instead model a beam of origin
$\vect{o}_s$ and direction $\vect{b}_s$ fixed to the tool, hitting a plane
$\{\vect{x}:\vect{m}^{\top}\vect{x}=h\}$ fixed in $W$, and take the range along
the beam,
\begin{equation}
\rho=\frac{h-\vect{m}^{\top}\!\left(\vect{p}+\R\,\vect{o}_s\right)}
{\vect{m}^{\top}\R\,\vect{b}_s},
\label{eq:rayplane}
\end{equation}
with eight parameters and unit scale. The range now depends on the whole
orientation: the fitted beam is 55.8$^\circ$ off the shaft axis and the plane
13.4$^\circ$ off the neutral axis, so tilting in opposite directions changes
$\rho$ in opposite senses, and absorbing that into a scale factor increases the
tip error (Sec.~\ref{sec:rotation}).
Because~\eqref{eq:rayplane} depends on $\R$, ToF also informs orientation, and
relies on vision for heading.

\subsection{Error-State Kalman Filter}
\label{sec:eskf}
An error-state Kalman filter splits the state in two: a nominal state
integrated openly from the IMU, and a small error state that the filter
estimates, injects into the nominal state and resets to zero after each
update~\cite{sola2017quaternion}. We use it because the orientation error is a minimal three-vector with no
quaternion constraint, and because the nominal state can be propagated at the
gyroscope's rate while late vision is applied whenever it arrives, as the
out-of-sequence update of Sec.~\ref{sec:rt} requires.

For the translation and rotation trials the nominal state is
$(\R,\vect{b}_g,L)$, with $\vect{b}_g$ the gyroscope bias, and the error state is
$\delta\vect{x}_s=[\delta\boldsymbol{\theta}^{\top},\delta\vect{b}_g^{\top},\delta
L]^{\top}\in\mathbb{R}^7$. The gyroscope propagates
$\R\leftarrow\R\,\mathrm{Exp}\big((\boldsymbol{\omega}-\vect{b}_g)\Delta t\big)$
through the IMU-to-tool mounting map. Updates are the accelerometer direction $\R^{\top}\vect{g}$
while $\|\boldsymbol{\omega}\|<2^\circ$/s, the two
constraints~\eqref{eq:plane} with the vision depth, and the ToF range
through~\eqref{eq:rayplane}. The tip follows from~\eqref{eq:tip} with $\vect{r}$
fixed, each stream shifted by its measured latency (Sec.~\ref{sec:latency}), and
the initial heading is picked among the branches satisfying~\eqref{eq:plane} by
reprojecting the shaft into the image.

\emph{Noise model.} Accelerometer, ToF and vision errors proved to be dominated
by pose-locked systematic components rather than white noise. The ToF depth,
for instance, scattered 0.43\,mm over repetitions of one pose but 3.04\,mm
across poses. Treating them as independent per sample made the filter over-confident. Each
$\sigma$ is
therefore set to the held-out residual, inflated by $\sqrt{N_c}$ for $N_c$
correlated samples per pose, every sample is used once, and the gyroscope-bias
prior is the measured bias instability of 0.05$^\circ$/s.

\subsection{Real-Time Tracking and Display}
\label{sec:rt}
Continuous motion leaves no stationary windows, so the tracking filter carries
the tip position and velocity as well,
$\delta\vect{x}_d=[\delta\vect{p}^{\top},\delta\vect{v}^{\top},
\delta\boldsymbol{\theta}^{\top},\delta\vect{b}_g^{\top},\delta\vect{b}_a^{\top},
\delta b_\rho]^{\top}\in\mathbb{R}^{16}$, with $\vect{b}_a$ the accelerometer
bias and $b_\rho$ a ToF range offset, propagated by the gyroscope and the raw
accelerometer. Because the IMU sits well proximal to the tip, its specific
force $\vect{a}_I$ is transferred there using the angular velocity
$\boldsymbol{\omega}$, the angular acceleration $\boldsymbol{\alpha}$ and the
lever arm $\vect{r}_I$ from the IMU to the tip,
\begin{equation}
\vect{a}_{\mathrm{tip}}=\vect{a}_I+\boldsymbol{\alpha}\times\vect{r}_I
+\boldsymbol{\omega}\times(\boldsymbol{\omega}\times\vect{r}_I),
\label{eq:lever}
\end{equation}
so gravity updates remain usable during motion and are down-weighted rather than
gated when $\|\vect{a}\|$ departs from $g$. The RCM enters as a soft
two-dimensional pseudo-measurement and ToF observes
$\|\vect{p}-\vect{r}\|+b_\rho$; depth and offset are indistinguishable from ToF
alone, so $b_\rho$ is estimated only while vision has supplied an independent
tip position within the last 2\,s. Vision contributes the shaft direction and
the tip position, the latter with larger variance along the line of sight, and a
frame is trusted only if its shaft line passes within 3.5\,mm of the registered
RCM (\emph{trocar gate}), since a line that misses it comes from a mis-segmented
frame or the wrong heading branch.

\begin{figure}[tb]
\centering
\includegraphics[width=0.86\columnwidth]{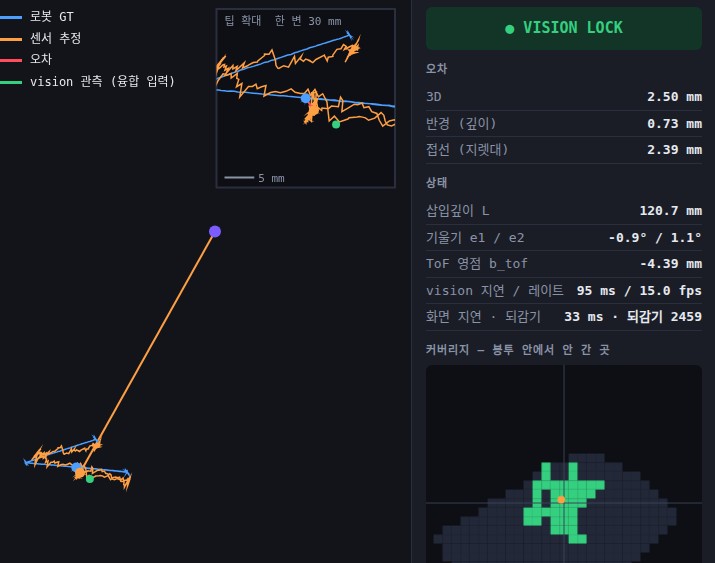}
\caption{The live display during a teleoperated block, at the gyroscope rate.
\emph{Left:} the tip path with the current pose (estimate orange, robot blue,
accepted vision observations green), inset a 30\,mm view around the tip.
\emph{Right, from the top:} the vision lock indicator; the error against the
robot, in three dimensions and split into radial and tangential parts; the
state --- insertion depth $L$, tilt, ToF offset $b_\rho$, vision latency and
display lag; and the working envelope visited so far.
}
\label{fig:gui}
\end{figure}

The same filter drives a display (Fig.~\ref{fig:gui}) that refreshes at the
gyroscope rate. Segmentation and reconstruction take about 113\,ms per frame, so
a vision observation describes the past: the filter rewinds to the frame's
capture time, applies the observation there and re-propagates. Its estimate is held 30\,ms behind real time rather
than 113\,ms, which the display reports as a 33\,ms end-to-end lag. Once all
inputs have arrived the live trajectory equals the offline causal filter, which
the dynamic results use without smoothing.

\section{Experiments}
\label{sec:experiments}
All experiments are referenced to an FR3 that holds the instrument through the
trocar of the phantom (Fig.~\ref{fig:overview}b), and none of them lets robot
data into the estimator: we evaluate static stability, observation latency,
translation, rotation and continuous tracking, with sensor ablations throughout.

\subsection{Protocol and Calibration}
Motions are commanded as minimum-jerk profiles about the RCM. Except for the
dynamic experiment, each trial brackets the motion with 3\,s stationary windows,
$W_{\mathrm{pre}}$ and $W_{\mathrm{post}}$, and the metric is the difference
between them, so constant offsets cancel. Trial order is randomized, and robot
ground truth is orders of magnitude finer than the sensor errors (0.0001$^\circ$
rotation repeatability, under 2\,$\mu$m of tip motion in pure roll). An ArUco marker was fixed to the instrument
\emph{for calibration only}, to command motions in the camera frame: the
hand-eye transform was solved in closed form~\cite{park1994axxb} and refined
over 30 poses, leaving 2.35\,mm and 0.74$^\circ$ (mean) on 20 held-out poses, to
which displacement metrics are insensitive.

\subsection{Static Stability}
\begin{table}[tb]
\caption{Static noise of the raw sensor streams}
\label{tab:static}
\centering\footnotesize
\input{tables/tab_static}
\end{table}
Twenty 30\,s holds at four poses and a 5\,min soak (Table~\ref{tab:static}) put
the accelerometer tilt noise at 0.30--0.55$^\circ$ and the raw ToF depth noise
at 2.95--3.43\,mm per sample. Both are the logger's own, since the robot tip
held to 0.019--0.023\,mm, and tilt drifted only 0.017$^\circ$ over 5\,min.

\subsection{Observation Latency}
\label{sec:latency}
\begin{table}[tb]
\caption{Observation latency of each channel}
\label{tab:latency}
\centering\footnotesize
\input{tables/tab_latency}
\end{table}
The robot excited insertion depth at 0.1--1.0\,Hz and shaft tilt and roll at
0.15--1.6\,Hz, with amplitudes bounded by a 50\,mm/s tip-speed limit. Lock-in
detection gave gain and phase against ground truth at each frequency, and we
fitted $G(j\omega)=K e^{-j\omega L_d}/(1+j\omega T)$. Raw ToF lagged by 9.0\,ms
against 58.7\,ms for the firmware-smoothed depth, consistent with the 39\,ms
moving average plus a zero-order hold, and the IMU and the camera lagged by
6.6--6.9 and 15.5--17.3\,ms (Table~\ref{tab:latency}). Camera latency was measured with a temporary fiducial but applies to the
markerless pipeline as well, since frames are time-stamped on arrival, before
any image processing.

\subsection{Translation Accuracy}
\begin{table}[tb]
\caption{Translation accuracy (300 trials) and sensor ablation}
\label{tab:translation}
\centering\footnotesize
\input{tables/tab_translation}
\end{table}

The tool, inserted 130\,mm, was moved along $\pm x,\pm y,\pm z$ of the camera
frame by 1, 2, 5, 10 and 20\,mm, ten times each (300 trials). This session
predates the segmentation network and the ray--plane range model: the shaft was
annotated by hand and, with the tilt below 9$^\circ$, the affine range map
sufficed. Constants were fitted on the 5, 10 and 20\,mm conditions only.

The displacement error was 1.21\,mm RMS, rising from 0.52\,mm at the held-out
1\,mm step to 2.12\,mm at 20\,mm (Table~\ref{tab:translation},
Fig.~\ref{fig:results}a). The scale was 0.962 and the error was mostly lateral.
Motion along $y$ was the weakest, 1.83\,mm against 0.61--0.82\,mm for $x$ and
$z$, because the constraint~\eqref{eq:plane} bounds only the image-normal
component. The ablations show the intended division of labor (Table~\ref{tab:translation}):
without vision the error grew tenfold and scaled with distance, because nothing
fixes heading; without ToF the direction survived but the scale collapsed to
0.749; and without accelerometer updates an error floor appeared even at the
1\,mm step.

\subsection{Rotation Accuracy}
\label{sec:rotation}
\begin{table}[tb]
\caption{Rotation accuracy (180 trials) and sensor ablation}
\label{tab:rotation}
\centering\footnotesize
\input{tables/tab_rotation}
\end{table}
With the tool inserted 140\,mm, the robot pitched and yawed the shaft about
axes perpendicular to its neutral direction (pitch $\pm5,\pm15,\pm25^\circ$;
yaw $-5,-8,-11,+5,+15,+25^\circ$, bounded by joint margin) and rolled it by
$\pm15,\pm45,\pm90^\circ$, ten times each (180 trials, all valid). The
labeler's tangent-plane residual here was 0.113$^\circ$ held out, against
0.335$^\circ$ for the hand-labeled translation session, and its junction depth
deviated from the robot by 0.16\,mm (median). Constants
were fitted on repetitions 1--3 and evaluated on 4--10, and we compare the
relative rotation $\R_{\mathrm{post}}\R_{\mathrm{pre}}^{\top}$ between the two
stationary windows with ground truth, \emph{without} any alignment.

\begin{figure}[tb]
\centering
\includegraphics[width=\columnwidth]{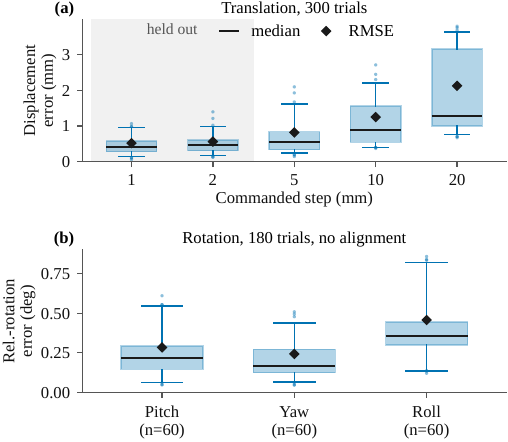}
\caption{Accuracy with stationary windows against the FR3 (median, quartiles,
5th--95th percentile whiskers, outliers). (a) Displacement error per commanded
step, 60 trials each; shaded steps were held out of calibration.
(b) Relative-rotation error per axis, no alignment. Diamonds mark the RMSE.}
\label{fig:results}
\end{figure}

The relative-rotation error was 0.341$^\circ$ RMS and the rotation-angle error
0.200$^\circ$ RMS, with a gain of 0.9997 from 5$^\circ$ to 90$^\circ$
(Table~\ref{tab:rotation}, Fig.~\ref{fig:results}b): 5.4$\times$ lower than the
BNO085 on-chip nine-axis orientation and 5.9$\times$ lower than gyroscope
integration. The tip displacement error was 0.76\,mm RMS,
and the insertion depth, constant at 140.0\,mm, was estimated at 140.00\,mm.
Without vision, heading is unobservable: the estimate is off by a 177$^\circ$
rotation about gravity and the relative-rotation error becomes 39.4$^\circ$.
Vision depth alone and ToF with vision heading gave 0.71 and 0.82\,mm, two
independent routes to the same depth. The worst condition was roll
$\pm90^\circ$, at 0.659$^\circ$, where the silhouette is blind to roll and the
gravity sensitivity is only $\sin 30.4^\circ$ for the neutral shaft of this
session.

\subsection{Dynamic Trajectory Tracking}
\label{sec:dynamic}
\begin{table}[tb]
\caption{Dynamic tip trajectory error, on programmed and teleoperated
continuous motion}
\label{tab:dynamic}
\centering\footnotesize
\input{tables/tab_dynamic_final}
\end{table}
\begin{figure*}[tb]
\centering
\includegraphics[width=\textwidth]{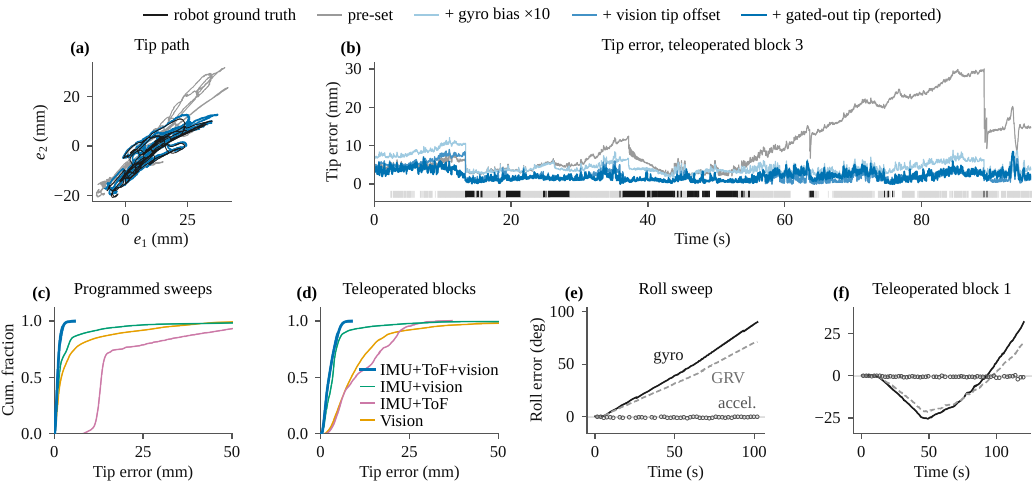}
\caption{Tracking of continuous instrument motion. (a) Tip path in teleoperated block~3 on the
plane normal to the neutral shaft and (b) its error over time at each rung of the correction
ladder of Table~\ref{tab:ladder}; ticks mark vision frames that passed the trocar gate (black) or
gave only their tip position (gray). (c,\,d) Tip error per sensor combination, all with the RCM.
(e,\,f) Roll error about the shaft from gyroscope integration (solid), the on-chip game rotation
vector (GRV, dashed) and the accelerometer (circles).}
\label{fig:dynamic}
\end{figure*}
With the tool at a working depth of 140\,mm, a fit-only block of ten static
poses (78\,s) calibrated the raw accelerometer and mounting, the vision tip
offset and the channel delays. Three programmed chirp sweeps followed,
102--137\,s each, with amplitudes decreasing with frequency under the 50\,mm/s
limit: pitch and roll over 0.15--1.6\,Hz (24.7 down to 3.0\,mm of tip travel,
25 down to 3.4$^\circ$) and depth over 0.1--1.0\,Hz (10 down to 6.4\,mm). An operator then drove four 120\,s free
trajectories, which we call \emph{teleoperated} blocks: a haptic stylus
(Touch, 3D Systems) moved by hand supplied the absolute tip targets that the RCM
servo of Sec.~\ref{sec:geom} executed (Fig.~\ref{fig:overview}c). Watching the live display, the operator kept the
depth within 130--150\,mm (85--99\% of the time) and the tilt within
10$^\circ$ (93--97\%), the validated range of monocular reconstruction, and the
true shaft line passed the registered RCM within 0.03\,mm (median). We score
the absolute tip error at every IMU sample in the robot base frame, without
alignment, excluding the first 20\% of each block as filter convergence
(Table~\ref{tab:dynamic}, Fig.~\ref{fig:dynamic}a,\,b).

The full fusion tracked the programmed sweeps with 1.22\,mm RMS tip error and
the teleoperated motion with 3.04\,mm, split between radial and tangential
directions as Table~\ref{tab:dynamic} shows, with 0.6--2.2$^\circ$ RMS
orientation error.

\emph{Sensor combinations.} On the same recordings with sensor subsets, all with
the RCM (Fig.~\ref{fig:dynamic}c,\,d), vision alone reached 13.4 and 27.5\,mm,
because nothing bridges the frames it rejects; 64--78\% of frames were accepted
and 14--44\% passed the trocar gate. IMU and ToF without vision drifted in
heading, IMU and vision without ToF lost depth, and only the full fusion stayed
at 1.22 and 3.04\,mm, the best combination in all seven blocks.

\emph{Degraded vision.} In an operating room the view is lost regularly, to
smoke, a soiled lens or the instrument leaving the field. On the same recordings
we kept one accepted frame in three, multiplied the vision axis noise fivefold,
and blanked vision for 1, 3 and 10\,s at sixteen points; the block-mean error
increased only from 1.22 to at most 1.35\,mm (programmed) and from 2.95 to at
most 3.24\,mm (teleoperated). Hiding the shaft junction alone, which removes the
independent tip and freezes the ToF offset, raised it to 6.2 and 7.7\,mm.

\begin{table}[tb]
\caption{Correction ladder: the three post-hoc changes, applied cumulatively}
\label{tab:ladder}
\centering\footnotesize\setlength{\tabcolsep}{2pt}
\input{tables/tab_ladder}
\end{table}
\emph{Gyroscope bias and post-hoc tuning.} The parameters fixed before
recording, which drove the live display, gave 3.45 and 10.28\,mm, almost
entirely tangential. Integrating the gyroscope about the shaft against ground
truth showed why (Fig.~\ref{fig:dynamic}e,\,f): its bias moved during motion and
accumulated 89$^\circ$ over the 102\,s roll sweep, while accelerometer-derived
roll stayed within 2.1$^\circ$. The neutral shaft of this session is
28.6$^\circ$ from vertical, so most of that error lies about gravity, where only
vision observes it. The results above re-run the same
recordings with three changes, which Table~\ref{tab:ladder} separates. Loosening
the bias alone lets the filter follow the drift but leaves the position to
vision: the sweeps improve slightly while the teleoperated blocks worsen.
Subtracting the 4\,mm vision tip offset fitted on the fit-only block fixes the
sweeps but not free motion, where the trocar gate accepts too few frames. Only
when the tip position, but not the axis, of gated-out frames is restored does
the teleoperated error fall to 3.04\,mm. The third change is thus required: the
loosened bias is safe only once vision constrains the position often enough. Pooled over all seven blocks the same loosening
worsened the vision-free IMU--ToF combination from 18.4 to 19.1\,mm, which is
therefore reported with the pre-set parameters. The choice of factor is not
critical: the block-mean teleoperated error was flat between 10$\times$ and
100$\times$ (2.87--2.99\,mm). Because these changes were
selected after inspecting the evaluation blocks, they remain to be confirmed on
an independently recorded session.

\section{Discussion}
The ablations on continuous motion show how the sensors complement each other,
and the magnetometer is the only one whose errors originate in the environment
rather than in the device: a nine-axis module carries a bench calibration into
an environment that invalidates it, whereas steel furniture and electrosurgical
equipment affect neither gravity, the range to a surface, nor the shaft
silhouette in an image. No magnetic quantity enters the accuracy reported here,
so the disturbance that limits nine-axis fusion outside the laboratory does not
reach it.

This has two implications. First, the approach could make routine recording
practical: nothing is added inside the patient, no marker is placed, the camera
already exists in the procedure, and the estimator never needs a robot, so a
case can leave behind the metric tip trajectory that imitation learning for
surgical robots needs, rather than a video from which pose must later be
estimated. Second, the estimate is causal and therefore available
\emph{during} the operation: a surgeon cannot see insertion depth or attitude
once the instrument is inside the patient, and the display provides both at the
gyroscope rate with a 30\,ms lag, so one device can serve for recording and for
intraoperative monitoring.

The study has limits. Results come from one device on a phantom with a fixed
ceiling camera, whereas an endoscope-mounted camera moves and adds its own pose,
and the free trajectories were teleoperated rather than hand-held so that robot
kinematics could serve as ground truth. Monocular reconstruction was
validated only near 140\,mm insertion, the RCM is assumed fixed, and the labeler
is tuned to one shaft appearance.

\section{Conclusion}
We set out to make hand-held laparoscopic instrument motion recordable under
operating-room constraints, where the part inside the patient cannot be modified
and the magnetic field cannot be trusted. A clip-on IMU and ToF rangefinder,
fused under the trocar RCM constraint with a markerless camera that replaces the
magnetometer as the heading reference, achieved this without robot data in the
estimator, with millimeter tracking accuracy, sub-degree relative rotation, and
the full fusion ahead of every sensor subset. Because no part of that estimate is magnetic, the method is suited to
operating-room use, where it could record expert motion across cases while
displaying depth and attitude live; validation there remains future work.

\let\oldbibliography\thebibliography
\renewcommand{\thebibliography}[1]{\oldbibliography{#1}\setlength{\itemsep}{0pt}\setlength{\parsep}{0pt}}
\bibliographystyle{IEEEtran}
\bibliography{refs_cr}

\end{document}

%% file: tables/tab_sensors.tex
\begin{tabular}{@{}>{\raggedright\arraybackslash}p{0.20\columnwidth}>{\raggedright\arraybackslash}p{0.195\columnwidth}>{\raggedright\arraybackslash}p{0.295\columnwidth}>{\raggedright\arraybackslash}p{0.15\columnwidth}@{}}
\toprule
Sensor & Measurement & Observes & Residual \\
\midrule
Gyroscope & Angular rate & All three rotations; drifts, bias estimated online & 0.05$^\circ$/s bias \\[1pt]
Accelerometer & Gravity $\R^{\top}\vect{g}$ & Shaft tilt and roll & 0.21$^\circ$ \\[1pt]
Magnetometer & \emph{Not used} & \emph{Would be heading (Sec.~\ref{sec:six})} & \emph{---} \\[1pt]
Camera & Tangent planes \eqref{eq:plane}, depth \eqref{eq:p0} & Heading; supports tilt, depth, lateral tip & 0.11$^\circ$, 0.47\,mm \\[1pt]
ToF (raw) & Ray--plane range \eqref{eq:rayplane} & Insertion depth $L$; weakly, orientation & 0.60\,mm \\[1pt]
Trocar (RCM) & Shaft line through $\vect{r}$ & Tip across the shaft & 0.02--0.03\,mm$^{\dagger}$ \\[1pt]
\bottomrule
\end{tabular}
\\[1pt]\parbox{\columnwidth}{\footnotesize RMS residual on held-out rotation repetitions 4--10; for the gyroscope, its bias instability. $^\dagger$Ground-truth shaft line to the RCM.}

%% file: tables/tab_static.tex
\begin{tabular}{@{}lccc@{}}
\toprule
Pose & Accel.\ tilt (deg) & Raw ToF depth (mm) & Robot tip (mm) \\
\midrule
P0 & 0.55$\pm$0.30 & 3.14$\pm$0.06 & 0.019$\pm$0.005 \\
P1 & 0.30$\pm$0.01 & 2.96$\pm$0.06 & 0.023$\pm$0.001 \\
P2 & 0.37$\pm$0.02 & 2.95$\pm$0.05 & 0.020$\pm$0.002 \\
P3 & 0.51$\pm$0.18 & 3.43$\pm$0.10 & 0.022$\pm$0.003 \\
\bottomrule
\end{tabular}
\\[1pt]\parbox{\columnwidth}{\footnotesize SD within each 30\,s hold, mean $\pm$ SD over the five holds; the last column is the robot's own motion.}

%% file: tables/tab_latency.tex
\begin{tabular}{@{}l@{\hspace{4pt}}cccc@{}}
\toprule
Channel & ToF raw & ToF smoothed & IMU & Camera \\
\midrule
Insertion depth & 9.0 (1.00) & 58.7 (1.00) & -- & 15.5 (1.03) \\
Shaft tilt & -- & -- & 6.9 (0.99) & 15.8 (1.02) \\
Roll & -- & -- & 6.6 (1.02) & 17.3 (1.00) \\
\bottomrule
\end{tabular}
\\[1pt]\parbox{\columnwidth}{\footnotesize Low-frequency group delay in ms, gain in parentheses. Smoothed ToF: the firmware 5-sample average; IMU: its on-chip orientation output.}

%% file: tables/tab_translation.tex
\begin{tabular}{@{}lcccc@{}}
\toprule
Step (mm) & $n$ & RMSE [95\% CI] (mm) & Median & 95th pct. \\
\midrule
1$^\dagger$ & 60 & 0.52 [0.45, 0.59] & 0.41 & 0.96 \\
2$^\dagger$ & 60 & 0.56 [0.49, 0.64] & 0.46 & 0.99 \\
5 & 60 & 0.82 [0.68, 0.96] & 0.55 & 1.62 \\
10 & 60 & 1.25 [1.08, 1.42] & 0.88 & 2.21 \\
20 & 60 & 2.12 [1.83, 2.41] & 1.28 & 3.63 \\
\midrule
All & 300 & 1.21 [1.08, 1.34] & 0.65 & 3.15 \\
\bottomrule
\end{tabular}

\vspace{4pt}

\begin{tabular}{@{}lcccc@{}}
\toprule
Configuration & RMSE (mm) & Scale & 1\,mm & 20\,mm \\
\midrule
Full fusion & 1.21 & 0.962 & 0.52 & 2.12 \\
No accel.\ update & 1.69 & 0.980 & 1.22 & 2.51 \\
No ToF & 5.97 & 0.749 & 0.61 & 11.58 \\
No vision & 12.65 & 0.955 & 1.31 & 24.73 \\
\bottomrule
\end{tabular}
\\[1pt]\parbox{\columnwidth}{\footnotesize $^\dagger$Held out of calibration (constants fitted on 5, 10, 20\,mm). CI: bootstrap over trials (5000 resamples). Scale: slope of estimated vs.\ true displacement magnitude.}

%% file: tables/tab_rotation.tex
\begin{tabular}{@{}lccc@{}}
\toprule
Configuration & Angle ($^\circ$) & Rel.\ rot.\ ($^\circ$) & Tip (mm) \\
\midrule
Full fusion & 0.200 & 0.341 & 0.76 \\
Affine ToF model & 0.198 & 0.340 & 2.58 \\
No vision depth & 0.201 & 0.341 & 0.82 \\
No ToF (vision depth) & 0.198 & 0.340 & 0.71 \\
No accel.\ update & 0.285 & 0.746 & 1.50 \\
No vision & 1.087 & 39.4 & 6.40 \\
\midrule
On-chip 9-axis fusion & 1.086 & -- & -- \\
Gyro integration & 1.170 & -- & -- \\
\bottomrule
\end{tabular}
\\[1pt]\parbox{\columnwidth}{\footnotesize RMSE over 180 trials. Angle: rotation magnitude; Rel.\ rot.: full relative-rotation error with no alignment to the robot; Tip: tip displacement.}

%% file: tables/tab_dynamic_final.tex
\begin{tabular}{@{}lcccccc@{}}
\toprule
\multicolumn{7}{@{}l}{\emph{(a) Full fusion, tip error per block (mm)}} \\
Block & RMSE & Median & 95th pct. & Radial & Tang. & Pre-set \\
\midrule
\multicolumn{7}{@{}l}{Programmed sweeps} \\
\quad Pitch & 1.56 &1.24 & 2.76 & 1.32 & 0.83 & 3.60 \\
\quad Roll & 1.08 &0.80 & 2.06 & 0.93 & 0.56 & 3.61 \\
\quad Depth & 1.02 &0.70 & 1.96 & 0.95 & 0.36 & 3.20 \\
\quad \emph{Pooled} &\emph{1.22} & \emph{0.90} & \emph{2.30} & \emph{1.07} & \emph{0.59} & \emph{3.45} \\
\midrule
\multicolumn{7}{@{}l}{Teleoperated blocks} \\
\quad Block 1 & 3.38 &2.81 & 5.64 & 2.52 & 2.25 & 11.99 \\
\quad Block 2 & 1.87 &1.35 & 3.48 & 1.69 & 0.79 & 3.63 \\
\quad Block 3 & 2.68 &2.09 & 4.67 & 2.13 & 1.63 & 13.41 \\
\quad Block 4 & 3.86 &3.68 & 5.98 & 2.83 & 2.62 & 9.26 \\
\quad \emph{Pooled} &\emph{3.04} & \emph{2.35} & \emph{5.42} & \emph{2.33} & \emph{1.95} & \emph{10.28} \\
\bottomrule
\end{tabular}
\\[1pt]\parbox{\columnwidth}{\footnotesize (a) Absolute error in the robot base frame, no alignment, first 20\% of each block excluded. Radial: along the shaft; Tang.: perpendicular. Pre-set: the parameters that drove the live display (Table~\ref{tab:ladder}).}

\vspace{5pt}

\begin{tabular}{@{}lcc@{}}
\toprule
\multicolumn{3}{@{}l}{\emph{(b) Sensor combinations, all with the RCM (mm)}} \\
Sensors & Programmed & Teleoperated \\
\midrule
Vision & 13.36 [31.3] & 27.52 [32.3] \\
IMU+ToF$^\ddagger$ & 23.36 [53.1] & 13.93 [24.4] \\
IMU+vision & 11.13 [18.2] & 9.33 [14.4] \\
IMU+ToF+vision & 1.22 [2.3] & 3.04 [5.4] \\
\bottomrule
\end{tabular}
\\[1pt]\parbox{\columnwidth}{\footnotesize (b) The same recordings re-run with sensor subsets: pooled RMSE, 95th percentile in brackets. $^\ddagger$Pre-set parameters, this combination's best setting in all seven blocks.}

%% file: tables/tab_ladder.tex
\begin{tabular}{@{}lccccc@{}}
\toprule
 & \multicolumn{2}{c}{Programmed (mm)} & \multicolumn{3}{c}{Teleoperated (mm)} \\
\cmidrule(lr){2-3}\cmidrule(lr){4-6}
Parameters & RMSE & Tang. & RMSE & Tang. & Worst \\
\midrule
Pre-set (live display) & 3.45 & 3.19 & 10.28 & 10.03 & 13.4 (B3) \\
$+$ bias random walk $\times$10 & 3.16 & 2.88 & 13.35 & 13.15 & 23.7 (B4) \\
$+$ vision tip offset & 1.24 & 0.70 & 13.35 & 13.21 & 23.4 (B4) \\
\emph{$+$ gated-out tips} & \emph{1.22} & \emph{0.59} & \emph{3.04} & \emph{1.95} & \emph{3.9 (B4)} \\
\bottomrule
\end{tabular}
\\[1pt]\parbox{\columnwidth}{\footnotesize Pooled tip error; each row adds one change to the row above, the last being what Table~\ref{tab:dynamic} reports. Tang.: perpendicular to the shaft, where heading error shows; Worst: the worst teleoperated block. Fixing the channel delays at the last row's values changes no entry by more than 0.2\,mm.}